\pdfoutput=1

\documentclass[11pt]{article}

\usepackage{emnlp2021}

\usepackage{times}
\usepackage{latexsym}

\usepackage[T1]{fontenc}

\usepackage[utf8]{inputenc}

\usepackage{microtype}
\usepackage{booktabs}
\usepackage{graphicx}
\usepackage{amsmath}
\usepackage{multirow}
\usepackage{float}
\title{Moving on from OntoNotes: Coreference Resolution Model Transfer}

\author{Patrick Xia \\
  Johns Hopkins University \\\
  \texttt{paxia@cs.jhu.edu} \\\And
  Benjamin Van Durme \\
  Johns Hopkins University \\\
  \texttt{vandurme@cs.jhu.edu} \\}

\date{}

\begin{document}
\maketitle
\begin{abstract}
Academic neural models for coreference resolution (coref) are typically trained on a single dataset, OntoNotes, and model improvements are benchmarked on that same dataset. However, real-world applications of coref depend on the annotation guidelines and the domain of the target dataset, which often differ from those of OntoNotes. We aim to quantify transferability of coref models based on the number of annotated documents available in the target dataset. We examine eleven target datasets and find that continued training is consistently effective and especially beneficial when there are few target documents. We establish new benchmarks across several datasets, including state-of-the-art results on PreCo.
\end{abstract}

\section{Introduction}

Starting initially with neurally-learned features \cite{clark-manning-2016-deep, clark-manning-2016-improving}, end-to-end neural models for coreference resolution (coref) \cite{lee-etal-2017-end, lee-etal-2018-higher} have been developed and imbued with the benefits from contextualized language modeling \cite{joshi-etal-2019-bert, joshi-etal-2020-spanbert} and additional pretraining \cite{wu-etal-2020-corefqa}. At the same time, the number of parameters used in these models have increased, raising questions of overfitting our research to a specific dataset. Several studies show that fully-trained neural models on preexisting large datasets do not transfer well to new domains \cite{aktas-etal-2020-adapting, bamman-etal-2020-annotated, DBLP:conf/aaai/TimmapathiniNMS21}, and that rule-based baselines can still be superior \cite{poot-van-cranenburgh-2020-benchmark}. Further, while prior work has analyzed fully-trained models for mention pairs, like gender bias \cite{rudinger-etal-2018-gender,webster-etal-2018-mind, zhao-etal-2019-gender}, there has not been a comprehensive comparison analyzing transfer across datasets for document-level coref.

We bridge the current gap in understanding between the strength of pretrained models in contrast to the value of annotated target data, in light of the strong few-shot capabilities demonstrated by pretrained language models \cite{brown2020language, schick2020its}. While transfer in other NLP tasks have been studied more in-depth, transfer in coref has scarcely been examined despite recent models containing hundreds of millions of parameters. We investigate model transfer across datasets with continued training, in which a fully-trained model on a source dataset is further trained on a small number of target dataset examples \cite{sennrich-etal-2016-improving, khayrallah-etal-2018-regularized}.\footnote{We use \textit{continued training} to refer to full model adaptation, in contrast to \textit{finetuning} which is more strongly associated with encoders that are trained without supervision \cite{Hinton2006}.}

We contribute the first study of neural coref transfer, showing that continued training is effective on eleven datasets spanning different domains, annotation guidelines, and languages. We find
evidence that OntoNotes, a widely-used but license-requiring dataset for benchmarking coref, is no better at model transfer than the freely-available PreCo. We establish modern benchmarks on several understudied datasets, including state-of-the-art results on PreCo. Additionally, we analyze practical considerations regarding model selection, catastrophic forgetting, and parameter sharing.\footnote{Code and pretrained models are available at \url{https://nlp.jhu.edu/coref-transfer}.}

\begin{table*}[t]
    \centering
    \small
    \renewcommand{\arraystretch}{1.6}
    \begin{tabular}{p{1.25cm}p{9.5cm}p{4cm}}
    \toprule
         Dataset & Example & Comments\\
    \midrule
         OntoNotes (general) & Judging from the Americana in [[Haruki Murakami's]\textsubscript{1} "A Wild Sheep Chase" [Kodansha]\textsubscript{2}, 320 pages, \$18.95]\textsubscript{3}, baby boomers on both sides of the Pacific have a lot in common.  & \textit{Only coreferring mentions are marked (no singletons).} \\
         ARRAU (news) & Judging from [the Americana in [[Haruki Murakami's]\textsubscript{1} "A Wild Sheep Chase" [[Kodansha]\textsubscript{2}, [320 pages]\textsubscript{3}, [\$18.95]\textsubscript{4}]\textsubscript{5}]\textsubscript{6}]\textsubscript{7}, [baby boomers on [both sides of [the Pacific]\textsubscript{8}]\textsubscript{9}]\textsubscript{10} have [a lot in [common]\textsubscript{11}]\textsubscript{12}. & \textit{All mentions are marked, even if they are singletons.} \\
         PreCo (general) & [Writer]\textsubscript{1}: [Ralph Ellison]\textsubscript{1} [Novel]\textsubscript{2}: [Invisible Man]\textsubscript{2} \hspace{2cm} \break [Invisible Man]\textsubscript{2} is [[Ellison's]\textsubscript{1} best known work]\textsubscript{2}, most likely because [it]\textsubscript{2} was [the only novel [he]\textsubscript{1} ever published during [[his]\textsubscript{1} lifetime]\textsubscript{3}]\textsubscript{2} and because [it]\textsubscript{2} won [him]\textsubscript{1} [the National Book Award]\textsubscript{4} in [1953]\textsubscript{5}. & \textit{Singleton mentions are marked. Many documents contain the title as its own sentence.} \\
         LitBank (books) & And [Jo]\textsubscript{1} shook the blue army sock till the needles rattled like castanets, and [her]\textsubscript{1} ball bounded across [the room]\textsubscript{2}. & \textit{Only certain ACE categories are marked.} \\
         QBCoref (trivia) & [This author]\textsubscript{1} wrote [a play]\textsubscript{2} in which [the queen]\textsubscript{3} [Atossa]\textsubscript{3} and [the ghost of [Darius]\textsubscript{4}]\textsubscript{5} react to news of a military defeat; [that play]\textsubscript{2} is [the only classical tragedy on a contemporary, rather than mythical, subject]\textsubscript{2}. & \textit{All characters, authors, and works are annotated. Other mentions are ignored.}\\
    \bottomrule
    \end{tabular}
    \caption{These examples from different datasets illustrate the differences in annotation standards, specifically for what is markable as a mention. Mentions are bracketed and entity clusters are subscripted with the same number. }
    \label{tab:examples}
\end{table*}

\section{Coreference Resolution}

Entity coreference resolution is the task of finding clusters of mentions within a document that all refer to the same entity. It still remains a difficult challenge in NLP due to several factors like ambiguity \cite{poesio-artstein-2008-anaphoric} and dependence on real-world knowledge \cite{levasque2012wino}. 

There are several large annotated datasets for coreference resolution.  Annotation guidelines for coref differ across these  datasets based on the intended goals of the creators, resulting in differences in what is considered a mention, how to handle singleton clusters,\footnote{An entity cluster with only one mention.} and what types of links should be annotated. Despite such differences, OntoNotes 5.0~\cite{weischedel2013ontonotes} emerged as the most widely-used benchmark for the full task, and widely used public models are based on this dataset \cite{manning-etal-2014-stanford, gardner-etal-2018-allennlp}. \autoref{tab:examples} shows the differences between OntoNotes and a few other datasets considered in this work.

However, OntoNotes-based models may not always be appropriate. OntoNotes is a collection of several thousand documents across just seven genres from the 2000s (or earlier), and many datasets fall outside of the scope of those genres or time period. Unlike other datasets, singletons are not annotated. In modeling OntoNotes, genre and speaker features are needed to improve on the state-of-the-art, both of which are idiosyncrasies of the OntoNotes dataset. It is unclear how well these models transfer to a new, target dataset, especially if it is annotated and usable in (continued) training.

Prior work on domain adaptation for coref has focused on a single dataset and often with non-neural models. \citet{yang-etal-2012-domain} use an adaptive ensemble which adjusts members per document. Meanwhile, \citet{zhao-ng-2014-domain} use an active learning approach to adapt a feature-based coref model to be on par with one trained from scratch while using far less data. \citet{moosavi-strube-2018-using} study model generalization by including carefully selected linguistic features, aiming to improve out-of-the-box general performance. \citet{aktas-etal-2020-adapting} adapt a model to Twitter by retraining with a target-dependent subset of genres of OntoNotes. 

While these studies shed insight on single datasets, we aim to set broader expectations and guidelines on effectively using new data for model adaptation, both in terms of quantity and allocation of data between training and model selection.

\section{Methods}

In this section, we describe the method, model, datasets, and initialization methods used to investigate the effectiveness of continued training.

\subsection{Continued Training}

We adopt the formulation of continued training from \citet{Luong-Manning:iwslt15} where a model is first trained on a source dataset until convergence. This fully-trained model is then used to initialize a second model which is trained on a target dataset. 

This framework has been used for other tasks where annotation guidelines or domains shift significantly between datasets, like in syntactic parsing \cite{joshi-etal-2018-extending}, semantic parsing \cite{fan-etal-2017-transfer, lialin2021update} and neural machine translation \cite{Luong-Manning:iwslt15, khayrallah-etal-2018-regularized}. In addition, continued training can be staggered at different granularities \cite{gururangan-etal-2020-dont} or use mixed in-domain and out-of-domain data \cite{xu2021gradual}.

\subsection{Incremental Coreference Model} End-to-end models for coreference resolution broadly have four parts: a text encoder, a scorer for mention \textit{detection}, a scorer for mention pair \textit{linking}, and an algorithm for decoding clusters. The incremental coreference (\textsc{ICoref}) model \cite{xia-etal-2020-incremental} used in this work is a constant-memory adaptation of the end-to-end neural coreference resolution model \cite{lee-etal-2017-end} with improvements from subsequent work that incorporates stronger encoders \cite{joshi-etal-2019-bert, joshi-etal-2020-spanbert}. By creating explicit clusters and performing mention-cluster linking instead of mention-pair linking, \textsc{ICoref} naturally produces clusters from linking scores. This memory-efficient model is conceptually similar to other recent cluster-based models \cite{toshniwal-etal-2020-learning, yu-etal-2020-cluster}. This model was chosen because of its competitive performance against the line of end-to-end neural coreference resolution models \cite{joshi-etal-2019-bert} and memory efficiency, which allows for experiments on longer documents. 

However, \textsc{ICoref}, like the models before it, is designed around OntoNotes. As a result, we make minor modifications for compatibility with other datasets by ignoring \textit{genre}-specific embeddings and implementing an auxiliary objective for entity mention detection, similar to the one adopted by \citet{zhang-etal-2018-neural}. For completion, we reformulate the \textsc{ICoref} model to more precisely describe these minor changes in \autoref{sec:appendix:model}.

\begin{table}[t]
    \centering
    \small
    \begin{tabular}{crrrr}
    \toprule
     Dataset & Training & Dev & Test & \# Folds \\
    \midrule
     OntoNotes\textsuperscript{en} & 2,802 & 343 & 348 & - \\
     OntoNotes\textsuperscript{zh} & 1,810 & 252 & 218 & - \\
     OntoNotes\textsuperscript{ar} & 359 & 44 & 44 & - \\
     PreCo & 36,120 & 500 & 500 & - \\
     LitBank & 80 & 10 & 10 & 10 \\
     QBCoref & 240 & 80 & 80 & 5 \\
     ARRAU\textsuperscript{RST} & 335 & 18 & 60 & - \\
     SARA & 138 & 28 & 28 & 7 \\
     Semeval\textsuperscript{ca} & 829 & 142 & 167 & - \\
     Semeval\textsuperscript{es} & 875 & 140 & 168 & - \\
     Semeval\textsuperscript{it} & 80 & 17 & 46 & - \\
     Semeval\textsuperscript{nl} & 145 & 23 & 72 & - \\
    \bottomrule
    \end{tabular}
    \caption{Number of documents for each of the datasets considered in this work. For the smaller datasets, we perform $k$-fold cross-validation. }
    \label{tab:data:stats}
\end{table}

\subsection{Data}
\label{sec:data}

We explore a total of two source datasets and eleven target datasets, described in \autoref{tab:data:stats}. For smaller datasets, evaluation is performed via $k$-fold cross-validation, following the original authors. 

\textbf{OntoNotes} 5.0 \cite{weischedel2013ontonotes} is a dataset spanning several genres including telephone conversations, newswire, newsgroups, broadcast news, broadcast conversations, weblogs, and religious text. The dataset contains annotations of syntactic parse trees, named entities, semantic roles, and coreference. Notably, however, it does not annotate for singleton mentions, while it does link events. It also includes data in English (\textsuperscript{en}), Chinese (\textsuperscript{zh}), and Arabic (\textsuperscript{ar}), which we refer to using superscripts.

\textbf{PreCo} \cite{chen-etal-2018-preco} is a dataset consisting of reading comprehension passages used in test questions. The authors argue that because its vocabulary is smaller than that of OntoNotes, it is more controllable for studying train-test overlap. While they detail many ways in which their annotation scheme differs from OntoNotes, we note that they annotate singleton mentions and do not annotate events. Furthermore, this corpus is sufficiently large that it is possible to train a general-purpose coreference resolution model. Finally, because the official test set has not been released, we refer to the official ``dev'' set as our test set, and use a separate 500 training examples as our ``dev'' set.

\textbf{LitBank} \cite{bamman-etal-2020-annotated} is an annotated dataset of the first, on average, 2,000 words of 100 public-domain books. While they annotate singletons, they also limit their mentions only to those which can be assigned an ACE category.

\textbf{QBCoref} \cite{guha-etal-2015-removing} is a set of 400 quiz bowl\footnote{Quiz bowl is a trivia competition where passages give increasingly easier hints towards a common answer, such as a book title, author, location, etc.} literature questions that are annotated for coreference resolution. This dataset also includes singleton annotations, and it only considers a small set of mention types. The documents are short and dense with (nested) entity mentions, as well as terminology specific to literature questions.

\textbf{ARRAU} \cite{uryupina_artstein_bristot_cavicchio_delogu_rodriguez_poesio_2020} is the second release\footnote{\texttt{LDC2013T22}} of ARRAU, a corpus first created by \citet{poesio-artstein-2008-anaphoric} which spans several genres. The fine-grained annotations mark the explicit type of coreference, and the dataset also includes phenomena like singleton mentions and non-referential mentions. We only use the coarsest-grained coreference resolution of the \texttt{RST} subcorpus, which is a subset of the Penn Treebank (PTB) newswire documents, and therefore uses the same splits as PTB \cite{poesio-etal-2018-anaphora}. Thus, this dataset overlaps with OntoNotes, which also includes sections of PTB. However, we can use ARRAU to study \textit{annotation} transfer.

\textbf{SARA} v2 \cite{holzenberger-etal-2021-factoring} is a collection of legal statutes in which text spans identified as arguments of legal structures are also annotated for coreference. Each document is a single short legal statute, and so the overall number of clusters is low while many clusters are singletons. 

\textbf{SemEval} 2010 Task 1 \cite{recasens-etal-2010-semeval} is a dataset for multilingual coreference resolution for studying the portability of coref systems across languages. It consists of data in English (overlapping with OntoNotes), German, Spanish (\textsuperscript{es}), Catalan (\textsuperscript{ca}), Italian (\textsuperscript{it}), and Dutch (\textsuperscript{nl}). Due to dataset overlaps and licensing, we only use the latter four languages in this paper.

\begin{figure*}[t]
    \centering
    \includegraphics[width=\linewidth]{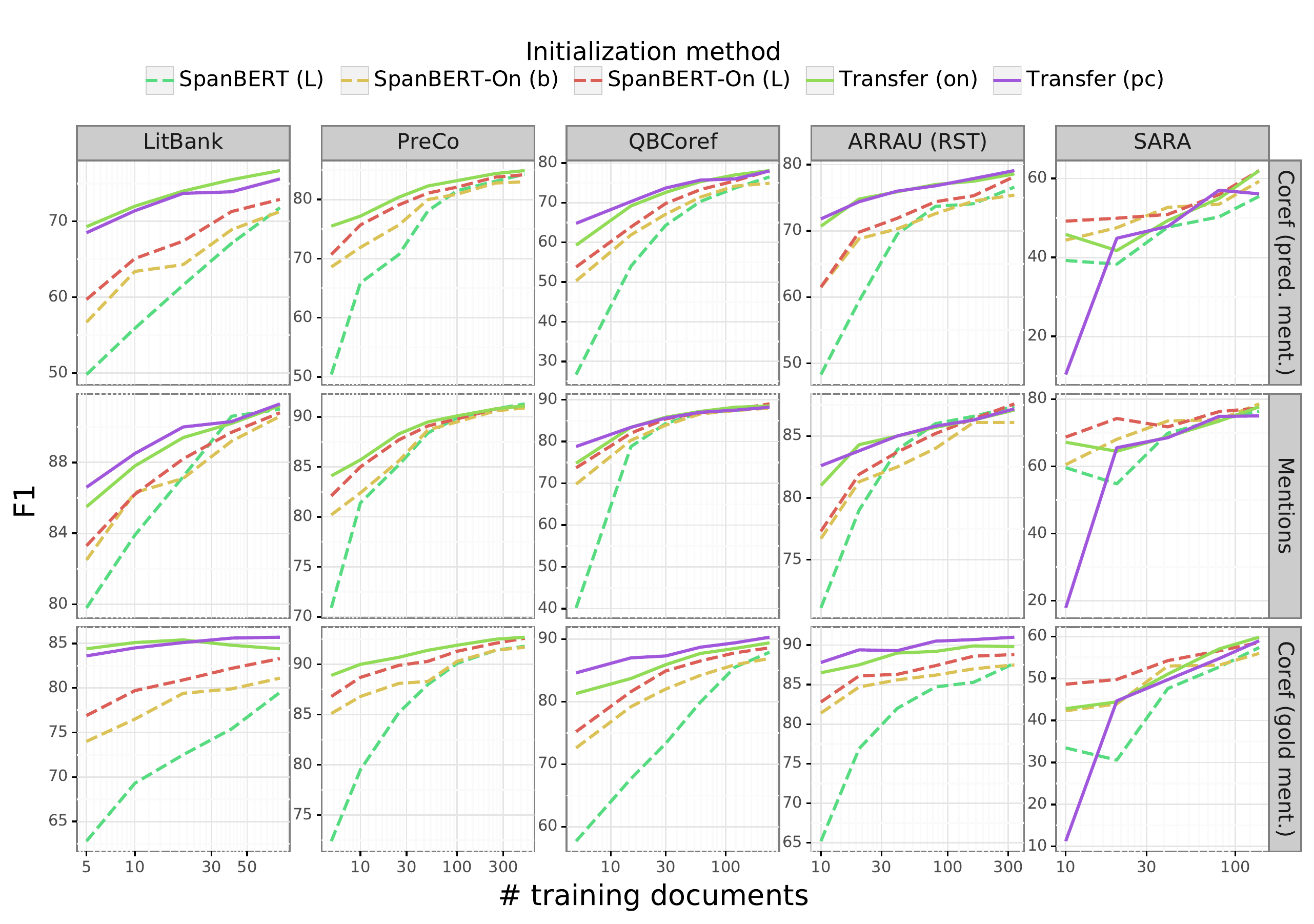}
    \caption{Each subplot shows the test performance for each model and (English) dataset when trained with a different number of documents. The first and second rows are coreference and mention boundary F\textsubscript{1} in the end-to-end setting, while the third row is the coreference F\textsubscript{1} with gold mentions. \textsc{SpanBERT} is a pretrained encoder, while the \textsc{SpanBERT-On} encoders are further finetuned on OntoNotes by \citet{joshi-etal-2020-spanbert}, with \textbf{b}ase and \textbf{L}arge designating its size. Unlike these (dashed lines) models for which we initialize the encoder, the \textsc{Transfer} models (solid lines) use continued training and initialize the full model with one that has already been trained on a source dataset, either OntoNotes (on) or PreCo (pc).}
    \label{fig:rq1links}
\end{figure*}

\subsection{Source models}
\label{sec:data:source}

\textsc{ICoref} has three trained components: an encoder, a mention scorer, and a mention linker. We explore initializing the encoder only and the full model.

\paragraph{Pretrained encoders} For these models, we initialize only the encoder with a pretrained one and randomly initialize the rest of the model. \citet{joshi-etal-2020-spanbert} trained the \textsc{SpanBERT} encoder on a collection of English data with a span boundary objective aimed at improving \textit{span} representations. In addition, they finetune \textsc{SpanBERT} by training a coreference resolution system on OntoNotes \cite{joshi-etal-2019-bert}, which they release separately. We name this finetuned encoder \textsc{SpanBERT-On}. \citet{conneau-etal-2020-unsupervised} trained \textsc{XLM-R}, a cross-lingual encoder, on webcrawled text in 100 languages. It is effective at cross-lingual transfer, including coreference linking \cite{xia2021lome}. We use the ``large" size of each model, except for one experiment with the ``base" size of \textsc{SpanBERT-on}.

\paragraph{Trained models} Alternatively, we can initialize with the full model. \textsc{Transfer (on)} is a model downloaded directly from \citet{xia-etal-2020-incremental}. We also train models on PreCo with SpanBERT-large (\textsc{Transfer (pc)}) and on OntoNotes\textsuperscript{en} with XLM-R (\textsc{Transfer (en)}).\footnote{We train the cross-lingual models separately because XLM-R and SpanBERT use different tokenization.} We also train a variant of each model with gold mention boundaries, which skips the mention scorer.

\section{Experiments and Results}

For a single source model and target dataset, we train several models using a different number of input training examples. The exact details for training set sizes and preprocessing are in \autoref{sec:appendix:dataset} while training details and hardware are in \autoref{sec:appendix:training}. We evaluate coreference using the average F\textsubscript{1} between MUC, B\textsuperscript{3} and CEAF\textsubscript{$\phi_4$}, following prior work \cite{pradhan-etal-2012-conll}.\footnote{We score exact match for SARA (following prior work).}

\subsection{How effective is continued training for domain adaptation?}
\label{sec:res:one}

\paragraph{Continued training} \autoref{fig:rq1links} shows that it is always beneficial to perform continued training on a source model, even if there is a large amount of target data. However, intuitively the differences are most pronounced in low-resource settings (with 10 fully-annotated documents) where it is still possible to adapt a strong model to perform non-randomly. These conclusions for coreference are similar to those drawn by \citet{gururangan-etal-2020-dont} on the effectiveness of domain- and task- pretraining of encoders for language classification tasks. These findings also support the intuition used by \citet{urbizu-etal-2020-sequence}, who choose PreCo as a pretraining corpus for ARRAU.

Continued training (and finetuning) is a core component of most NLP models, as text embeddings are typically derived from large pretrained models. \citet{joshi-etal-2018-extending} find that model adaptation with contextualized word embeddings only requires a small set of partial annotations in the new domain for syntactic parsing. Meanwhile, \citet{brown2020language} and \citet{schick2020its} find that pretrained language models can effectively learn a broad suite of sentence-level understanding, translation, and question-answering tasks with just a few examples. We corroborate their findings for a document-level information extraction task, since our models, based on strong pretrained encoders, perform well with just 5 or 10 training documents. 

\paragraph{OntoNotes vs. PreCo} We find that OntoNotes (\textsc{Transfer (on)}), despite being the benchmark dataset, is on par (or worse) as a pretraining dataset compared to PreCo (\textsc{Transfer (pc)}). One possibility is that because PreCo annotates for singletons, it is closer to the target datasets that also annotate singletons. This is evident when we compare the mention detection accuracy of the two models in low-data settings (e.g. LitBank or QBCoref at 5 examples). However, we subsequently explore the case when all models are given gold mention boundaries in pretraining, continued training, and testing, which would effectively evaluate just the linker. We find in this case that PreCo outperforms OntoNotes even more on QBCoref, LitBank, as well as ARRAU\textsuperscript{RST}. This suggests PreCo as a preferred pretraining dataset over OntoNotes when there are few annotated documents.

\paragraph{Model size and pretraining} The publicly available models use the ``base'' and ``large'' encoders. While there are even larger encoders, coreference models using them are rare. For future model development, one may decide between using a publicly available small model and retraining a large one from scratch. To simulate this, we compare a small encoder finetuned on OntoNotes, \textsc{SpanBERT-On (b)}, with \textsc{SpanBERT (L)}, which has not been trained on the task. This is also a realistic setting if there are hardware or compute limitations. 

In all datasets, we see that there is benefit to having some pretraining. When there is not much training data, the smaller (finetuned) encoder outperforms the larger encoder without finetuning. However, with enough data, the large model appears to surpass the smaller model. Nonetheless, there exist scenarios where continued training of a smaller model is desirable.

\begin{table*}
    \centering
    \setlength{\tabcolsep}{3pt}
    \small
    \begin{tabular}{cccccc}
    \toprule
    Dataset & Prior work & Previous Model & Previous Score & Our best & Our Model \\
    \midrule
     PreCo  & \citet{wu2020understanding}  & SpanBERT + C2F & 85.0 & {\bf 88.0} & \textsc{pc}\\
     LitBank & \citet{thirukovalluru-etal-2021-scaling} & SpanBERT + C2F & {\bf 78.4} & 76.7 & \textsc{on} \\
     QBCoref & \citet{guha-etal-2015-removing} & Berkeley & $<35$ & {\bf 78.1} & \textsc{on} \\
     ARRAU\textsuperscript{RST} &  \citet{yu-etal-2020-cluster} & BERT + cluster ranking & 77.9 & {\bf 79.1}* & \textsc{pc} \\
     SARA & \newcite{holzenberger-etal-2021-factoring} & string match baselines & 55.1 & {\bf 72.9} & \textsc{on} \\
     \midrule
     OntoNotes\textsuperscript{zh}  & \citet{chen-ng-2012-chinese} & Multi-pass sieve & 62.2 & {\bf 69.0} & \textsc{en}\\
     OntoNotes\textsuperscript{ar}  & \citet{aloraini-etal-2020-neural} & AraBERT + C2F & {\bf 63.9} & 58.5 & \textsc{en}\\
     SemEval\textsuperscript{ca}  & \citet{attardi-etal-2010-tanl} & feature-based + MaxEnt & 48.2 &  {\bf 51.0} & \textsc{en}\\
     SemEval\textsuperscript{es}  & \citet{attardi-etal-2010-tanl} & feature-based + MaxEnt & 49.0 &  {\bf 51.3} & \textsc{en}\\
     SemEval\textsuperscript{it}  &  \citet{kobdani-schutze-2010-sucre} & feature-based + decision tree & {\bf 60.8} & 36.7 & \textsc{en}\\
     SemEval\textsuperscript{nl}  & \citet{kobdani-schutze-2010-sucre} & feature-based + decision tree & 19.1  & {\bf 55.4} & \textsc{en}\\
    \bottomrule
    \end{tabular}
    \caption{Test F\textsubscript{1} on all datasets and the previous state-of-the-art on each dataset, to the best of our knowledge. Again, we are benchmarking the general method of continued training described in this paper, which will not necessarily outperform models that incorporate domain or language specific knowledge. Our best \textsc{Transfer} model is determined by the dev set (\autoref{sec:appendix:training}). *ARRAU\textsuperscript{RST} is not directly comparable to prior work as we test on a slightly differently-preprocessed subset. Multi-pass sieve \cite{raghunathan-etal-2010-multi}, Berkeley \cite{durrett-klein-2013-easy}, and C2F \cite{lee-etal-2018-higher} refer to widely-used coreference resolution models.}
    \label{tab:data:benchmark}
    \setlength{\tabcolsep}{6pt}
\end{table*}

\begin{figure*}[h]
    \centering
    \includegraphics[width=\linewidth]{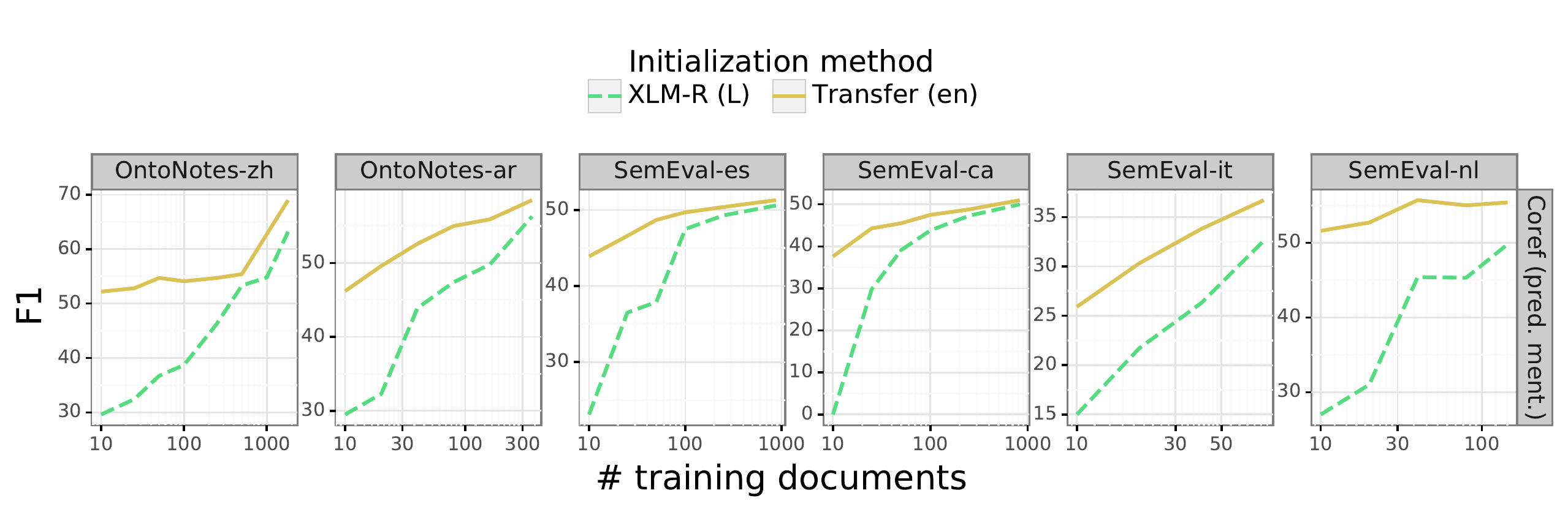}
    \caption{Like \autoref{fig:rq1links}, this plot demonstates the effectiveness of continued training across different \textit{languages}. \textsc{XLM-R} uses a pretrained encoder (dashed line), while \textsc{Transfer (en)} is first trained on OntoNotes\textsuperscript{en} (solid line). Trends on mention accuracy and using gold mentions look similar and are in \autoref{sec:appendix:layers}.}
    \label{fig:rq5links}
\end{figure*}

\paragraph{New benchmarks} \autoref{tab:data:benchmark} shows the test scores of our best model compared to prior work. For PreCo, we directly evaluate on the fully-trained model without continued training, as the full dataset is sufficiently large. Since some of these datasets are understudied, we present these as stronger baselines for future work.\footnote{Contemporaneous work has established even stronger baselines for LitBank \cite{thirukovalluru-etal-2021-scaling}.} The purpose is to quantify the effectiveness of continued training and highlight PreCo as an alternative pretraining dataset. %
Note that we achieve this strong performance without hyperparameter tuning or incorporating any language or domain specific features. %

\paragraph{Cross-lingual transfer} We present the results for multilingual coreference resolution in \autoref{fig:rq5links}. The gap in performance at low-data conditions (and the high initial starting point) shows that transfer via continued training is also effective cross-lingually in the end-to-end document-level task. Our results corroborate prior work \cite{conneau-etal-2020-unsupervised} by providing more evidence for XLM-R's cross-lingual transfer ability, in this case on the full end-to-end task. Given these results, we expect \textit{joint} multilingual pretraining followed by continued training to be an even more effective recipe in creating the best models for each language. This is out of scope for this work, which is focused on transfer from single datasets.

\begin{figure*}[t]
    \centering
    \includegraphics[width=\linewidth]{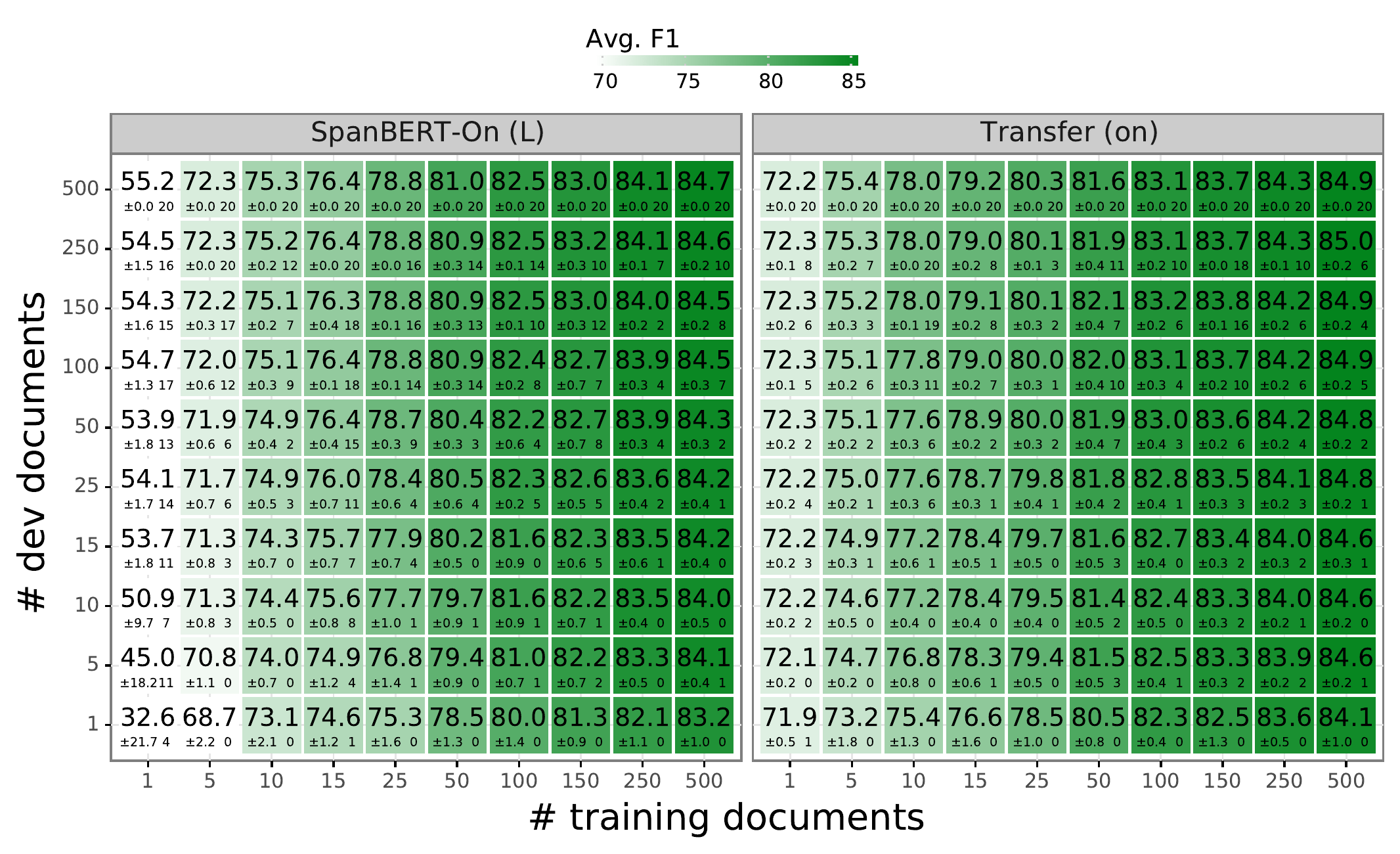}
    \caption{The expected test F\textsubscript{1} (and standard deviation) on the PreCo dataset for a given number of training documents and 20 sampled subsets of dev documents for two models described in Section \ref{sec:data:source}. The number of runs matching the best full dev checkpoint is in the lower-right. We find that the dev set size has relatively little impact.}
    \label{fig:rq2dev}
\end{figure*}

\begin{figure*}
    \centering
    \includegraphics[width=\linewidth]{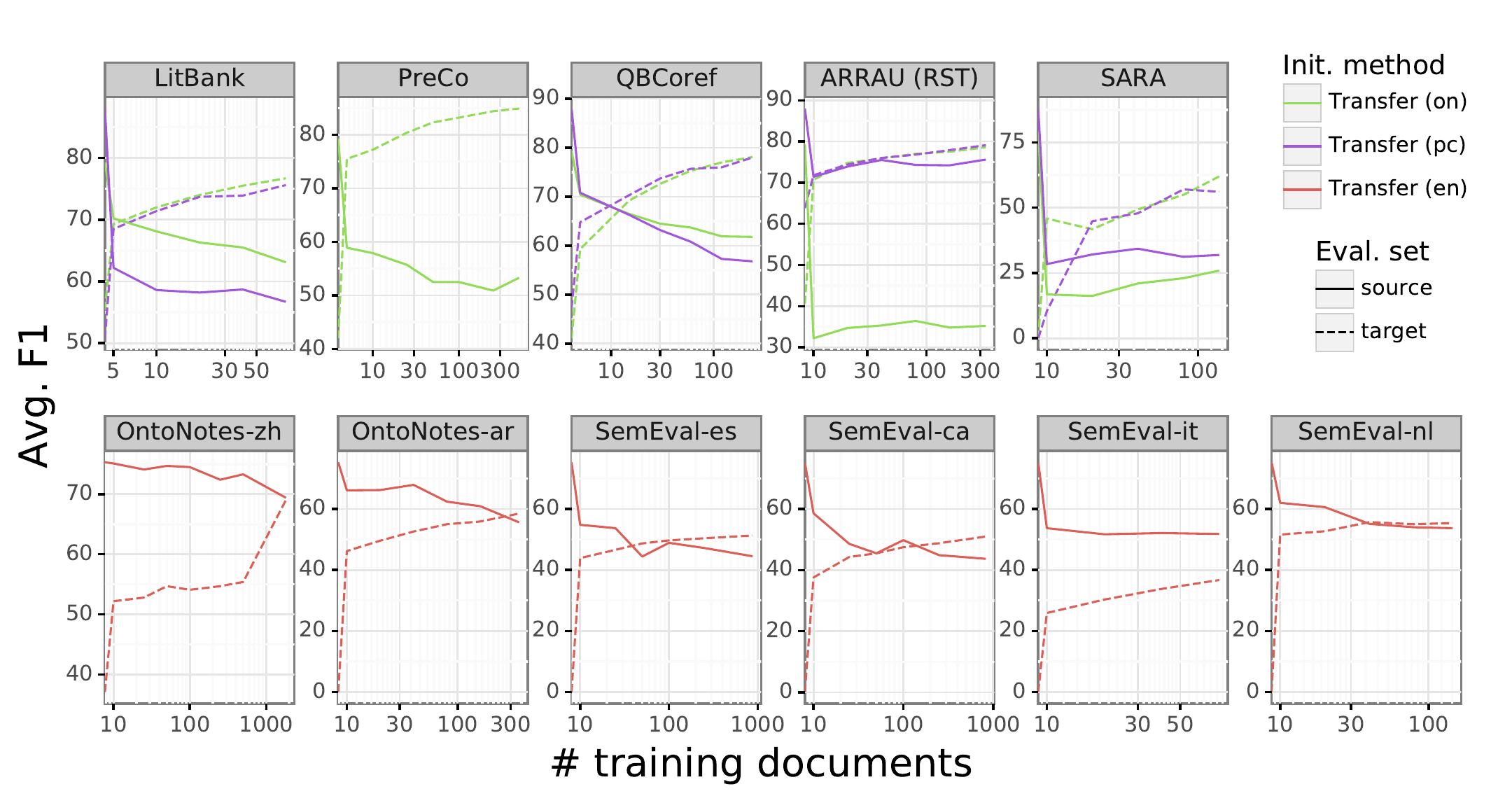}
    \caption{Average F\textsubscript{1} of the models on both the target and the original datasets as different number of (target) training examples are used in continued training. The dashed lines are the scores on the target dataset (mirroring \autoref{fig:rq1links}) while the solid lines show performance on the original dataset.}
    \label{fig:rq3}
\end{figure*}

\subsection{How to allocate annotated documents?}
\label{sec:res:alloc}

In \autoref{fig:rq1links}, the experiments for each dataset used the same dev set for model selection to improve comparability. At the same time, we observe that adding even a few more training examples can lead to improved performance. For some datasets, like PreCo, the size of the dev set used for model selection in our experiments greatly outnumbers the number of training documents. Here, we explore allocating fewer documents for model selection.

We compare 20 models for PreCo trained with a different number of examples using \textsc{SpanBERT-On (L)} and \textsc{Transfer (On)}. We train each model for 60 epochs and make predictions on all 500 dev examples. Next, for each dev set size, we sample a subset of the full predictions and determine, post-hoc, the checkpoint at which the model would have stopped had we used that sampled subset. We sample 20 such subsets and compute the expected scores and standard deviation for each model, along with how frequently the subset agreed with the full dev set.

\autoref{fig:rq2dev} summarizes the results, showing remarkable stability in expectation even with tiny dev sets, often less than a couple points behind using the full dev set. Given a fixed budget of documents or annotations, these results suggest that it is beneficial to allocate as many documents as possible towards training, leaving behind a small set for model selection.

\subsection{How much do the source models forget?}
\label{sec:res:forget}

 To measure the degree of catastrophic forgetting \cite{MCCLOSKEY1989109}, we revisit the source datasets of each \textsc{Transfer} model and track its performance in the presence of more training data.\footnote{For datasets with $k$-folds, we plot the mean across folds.} In \autoref{fig:rq3}, we see that on some datasets, the performance difference is especially pronounced after training on just 10 examples in the target dataset. 
 
 We hypothesize that this is due to easy-to-learn changes between the annotation guidelines that are incompatible between the two datasets, like the annotation of certain entity types. Two pairs, (OntoNotes\textsuperscript{en}$\rightarrow$OntoNotes\textsuperscript{zh}) and (PreCo$\rightarrow$ARRAU\textsuperscript{RST}) are less affected by continued training. For OntoNotes, the same guidelines are used for all languages. Meanwhile, PreCo and ARRAU\textsuperscript{RST} are more similar in annotation guidelines than any other pair since they both include singletons. On the other hand, (OntoNotes$\rightarrow$ARRAU\textsuperscript{RST}) shows a substantial drop in performance despite the two datasets containing overlapping documents. 
 
 In the cross-lingual setting, we observe that the drops are smaller than across English datasets. This could be due to several factors. The XLM-R encoder is already trained multilingually and has strong crosslingual performance \cite{conneau-etal-2020-unsupervised}, while English encoders are not well-suited for all domains, like law \cite{chalkidis-etal-2020-legal}. The crosslingual datasets in this study (OntoNotes and SemEval) are primarily in the same domain (newswire) and share similar annotation guidelines. And, in some cases where the trend looks flatter (SemEval\textsuperscript{it}, Semeval\textsuperscript{nl}, and even SARA), the training dataset is also smaller.

 \begin{figure}[H]
    \centering
    \includegraphics[width=\linewidth]{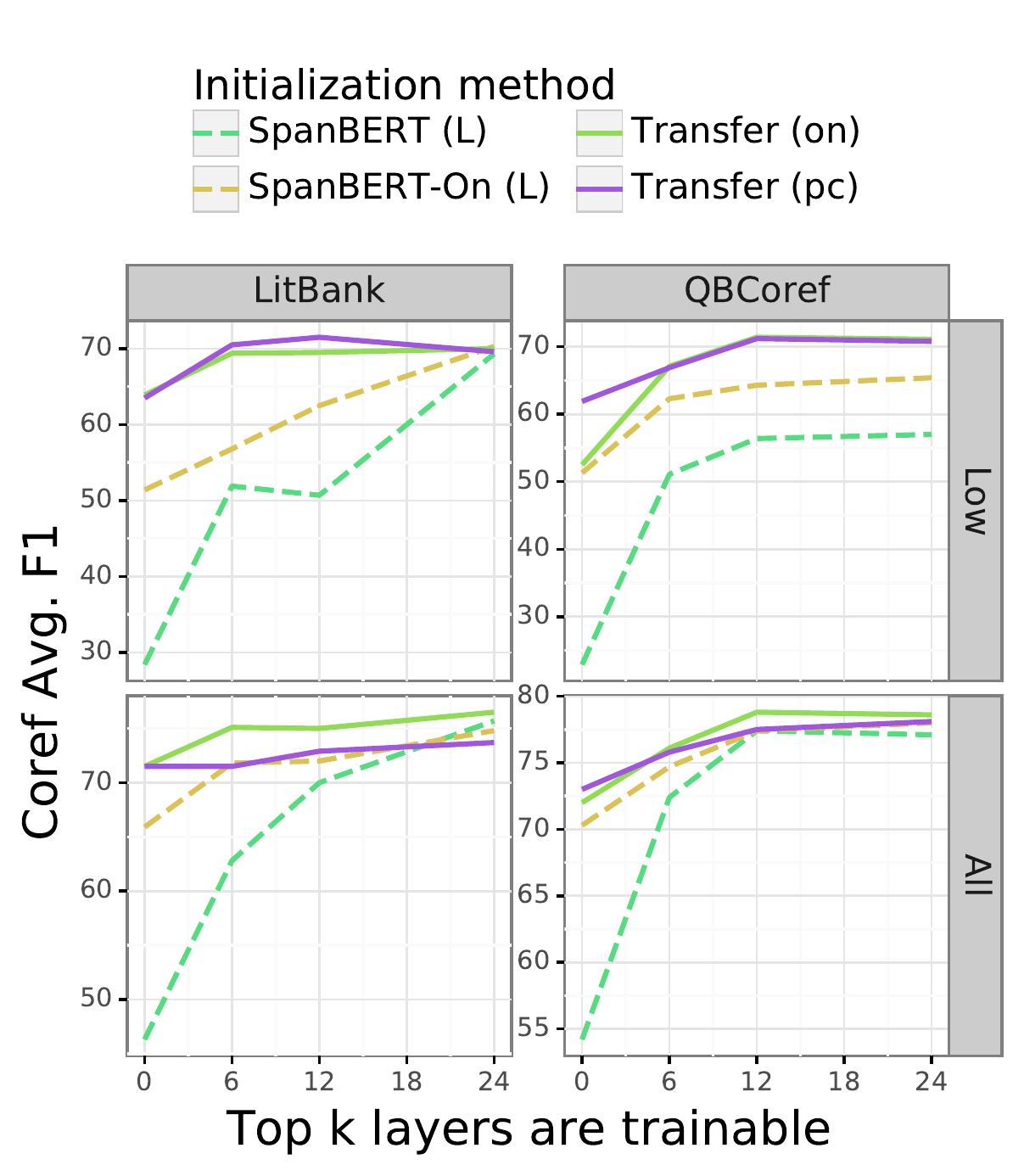}
    \caption{Average F\textsubscript{1} across different models and number of trainable layers. \textit{Low} vs. \textit{All} describes the number of documents used for the first fold of LitBank (10 vs. 80) and QBCoref (15 vs. 240).}
    \label{fig:rq4}
\end{figure}

 \subsection{Which encoder layers are important?}
 \label{sec:res:layers}
 
 Training the entire encoder is an expensive cost of continued training, both in terms of training time and in the number of \textit{new} parameters introduced by a new target dataset. We consider freezing some parameters of the encoder and training the top-$k$ layers, along with the rest of the model, for each of the ``large'' encoders. We investigate LitBank and QBCoref under low and high(er) data conditions. This is motivated by prior work which uses just the top four layers \cite{aloraini-etal-2020-neural} and by findings from encoder probing that higher layers are more salient for coreference \cite{tenney-etal-2019-bert}.

 \autoref{fig:rq4} shows that there are gains to training some layers, but it is not always necessary to train the full model. In particular, for transferred models, we observe that unfreezing more layers of the encoder could even lead to worse performance. On the other hand, untrained models generally benefit from training more of the encoder. These trends are observed in both datasets and data quantities.\footnote{This is also observed for OntoNotes\textsuperscript{zh} and in medium data conditions, detailed in \autoref{sec:appendix:layers}.}

 This demonstrates that continued training allows us to freeze a substantial fraction of the model and still achieve good performance. In a multi-dataset scenario, this would also reduce the total number of parameters as the lower layers of the encoder can be shared. This is impactful for neural coref models because recent improvements are due to encoders that are also growing in size (340M for SpanBERT (L) and 559M for XLM-R), which are significantly larger than the rest of the model (40M).

\section{Conclusion}

We comprehensively examine the transferability of neural coreference resolution models. We explore several model initialization methods across a wide set of domains and languages, and with a different number of training examples, to demonstrate the universal effectiveness of continued training. Additionally, this method results in improved performance over prior work on many of these datasets. Furthermore, we find that PreCo can be effectively used for pretraining, suggesting a viable alternative to OntoNotes.%

In our analysis, we find that: given a fixed number of annotated documents, few need to be allocated for model selection; continued training also suffers from catastrophic forgetting; and continued training is effective with partially frozen encoders. This study and its set of benchmarks serve as a reference for future work in coreference resolution model adaptation, especially for scenarios where annotation can be expensive or data may be scarce.

\section*{Acknowledgments}
We thank Huda Khayrallah, Shubham Toshniwal, and Michelle Yuan for guidance early in this work. We also thank Seth Ebner, members of JHU CLSP, and anonymous reviewers for helpful discussions and feedback. This work was supported in part by DARPA AIDA (FA8750-18-2-0015). The views and conclusions contained in this work are those of the authors and should not be interpreted as necessarily representing the official policies, either expressed or implied, or endorsements of DARPA or the U.S. Government. The U.S. Government is authorized to reproduce and distribute reprints for governmental purposes notwithstanding any copyright annotation therein.

\bibliographystyle{acl_natbib}
\bibliography{anthology,custom}

\clearpage
\newpage

\appendix

\section{Model}
\label{sec:appendix:model}

\subsection{The \textsc{ICoref} model}

The \textsc{ICoref} model \cite{xia-etal-2020-incremental} uses an incremental algorithm to perform coreference resolution. Given a text segment of length $n$ with (sub)tokens $x_1 \ldots x_n$, the model enumerates all spans $x_{a,b} \in X$, where $x_{a,b} = [x_a, x_{a+1}, ..., x_b]$ up to a certain length, respecting sentence boundaries. The span embedding $\mathbf{x}_{a:b}$ is then computed as a function of the component embeddings, determined by the output of an encoder: $\mathbf{x}_{a:b} = [\mathbf{x}_a; \mathbf{x}_b; f([\mathbf{x}_a, ..., \mathbf{x}_b]; \phi(a,b))]$ where $f$ is an attention-weighted average and $\phi(a,b)$ is a width feature. This is identical to the representation used by \citet{lee-etal-2017-end}. Like prior work, we learn a span scoring function $s_m(x_i)$ intended to rank the likelihood the given span is a coreference mention. We prune the number of spans considered in the next step to a manageable number spans, $kn$, for some ratio $k$.

The incremental algorithm iterates through the spans, collecting a list of clusters, $C$ (initially empty). Each span $x_i$ is scored by a pairwise scorer, $s_c(x_i, c)$, against the clusters already found by the model. Specifically, $s_c(x_i, c) = s_m(x_i) + s_a(x_i, c)$, which means this score is influenced by the likelihood $x_i$ is a mention. This is akin to the pairwise antecedent scorer from prior work. However, in \textsc{ICoref}, the scores are computed against clusters instead of against spans, which reduces the need for cluster decoding later.

If $\max_{c_j \in C}(s_c(x_i, c_j)) \leq 0$, a new cluster, $c_\text{new} = \{x_i\}$ with embedding $\mathbf{x}_i$, is created and added to $C$. Otherwise, $x_i$ is merged into the top-scoring $c_j$, with the new embedding, $$\mathbf{c}_j' = \alpha\mathbf{x}_i + (1-\alpha)\mathbf{c}_j,$$ where $\alpha$ is a learned function of $x_i$ and $c_j$.

The training objective aims to minimize $-\log \prod_{x_i \in X} P(c^{*}_{x_i} | x_i)$, where $c^{*}_{x_i}$ is the correct cluster determined by the cluster containing the most recent antecedent of $x_i$. If no such antecedent exists, then the correct cluster is the dummy cluster, $\epsilon$, and $s_c(x_i, \epsilon) = 0$. Letting $C_{\epsilon} = C \cup \{\epsilon\}$, the probability can then be computed as

$$P(c^{*}_{x_i} | x_i) = \frac{\exp(s_c(x_i, c^{*}_{x_i}))}{\sum_{c_j \in C_{\epsilon}} \exp(s_c(x_i, c_j))}.$$ 

In this work, we instead optimize for \textit{all} antecedents of $x$, $Ant(x)$, instead of the most recent one:

\begin{equation}
    -\log \prod_{x_i \in X} \sum_{y_i \in Ant(x)} \frac{1}{|Ant(x_i)|}P(c_{y_i} | x_i).
\end{equation}

We find that this leads to comparable (or slightly better) performance.

Finally, $s_a$ usually incorporates a genre embedding determined by the genre of the document. We retain that small set of parameters but assume all documents have the same genre. The only model for which this is not the case is the directly downloaded model, as it was trained for best performance on OntoNotes.

For most datasets and many downstream tasks, we want to include the singleton entity mentions in the output predictions. For OntoNotes, all singleton mentions are removed in postprocessing. We could add an auxiliary objective that maximizes $s_m(x_i)$ if $x_i$ is an entity mention \cite{zhang-etal-2018-neural} and only prune out singleton mentions $s_m(x_i) < 0$ in postprocessing. Instead, we present a model reformulation that is similar to the choices made by \citet{toshniwal-etal-2020-learning}.

Instead of taking the top $kn$ spans at span pruning, we prune to the top $kn$ spans from the set $\{x_i \in X: s_m(x_i) > 0\}$ (which could have fewer than $kn$ elements). This is both more efficient and easier to optimize for. Now, the training objective is to minimize $s_m(x_i)$ if $x_i$ is not an entity mention, and maximize $s_m(x_i) + s_a(x_i, c_j)$ if it is. This latter term is identical to $s_c(x_i)$ from the previous model.

We can interpret this change as now modeling the joint distribution of whether $x_i$ is an entity mention (a binary random variable $M$) and which entity cluster ($E$) is would best match to ($s_a$). We can decompose the joint probability, 
$$P(E, M \mid x_i) = \sum_{m \in \{0, 1\}}P(E|m, x_i)P(m, x_i).$$ 

This can further split into the components, 
\begin{align}
  P(E|M=1, x_i) &= \frac{\exp(s_a(x_i, c^{*}_{x_i}))}{\sum_{c_j \in C_\epsilon} \exp(s_a(x_i, c_j))} \\
  P(E|M=0, x_i) &= 1 \\
  P(M=1, x_i) &= \frac{\exp(s_m(x_i))}{1+\exp{s_m(x_i)}}\\
  P(M=0, x_i) &= 1 - P(M=1, x_i)
\end{align}

The $M=1$ objective is the same as training without singleton mentions (as in OntoNotes), while the $M=0$ term accounts for singletons. Note that if we know $M=0$, then we always make the correct ``cluster'' decision by ignoring it for the remainder of the algorithm, which allows for this simplification.

This is different from simply adding an objective maximizing $P(M)$, since that would incorrectly handle cases when $M=0$. In practice, however, we found that this makes no difference in performance on the task, though pruning spans earlier resulted in a substantially faster model.

\section{Dataset Preprocessing}
\label{sec:appendix:dataset}

We use the scripts from \citet{joshi-etal-2019-bert} to convert all documents into sentence-separated and subtokenized segments of sizes at most 512. For all English datasets, we use the SpanBERT tokenizer, while we use the XLM-R tokenizer for the cross-lingual experiments.

For QBCoref, we split the dataset into five splits after shuffling the initial dataset. For LitBank, we use the published splits \cite{bamman-etal-2020-annotated}. In ARRAU\textsuperscript{RST}, several mentions are split. Correctly modeling split spans is an active area of ongoing work \cite{yu-etal-2020-free, yu2021stay}. Since we use ARRAU\textsuperscript{RST} primarily for intrinsic comparisons, we defer to the \textit{minimum} span if a mention is split. This means we replaced a subset of markables, listed in \autoref{tab:appendix:arrau}. In addition, a small number of markables do not have an annotated coreference cluster, while a couple split markables failed to reduce because there is no minimum span annotated. These two phenomena did not affect the test set. Nonetheless, the model's inability to address split markables affects comparability against prior work.

\begin{table}
    \centering
    \small
    \begin{tabular}{ccccc}
    \toprule
    Split & Total & Split &  No ``coref'' & No ``min'' \\
    \midrule
     train & 57,686 & 677 & 4 & 2\\
     dev  & 3,986 & 40 & 0 & 0\\
     test & 10,341 & 145 & 0 & 0 \\
    \bottomrule
    \end{tabular}
    \caption{Statistics of markables that are either reduced or ignored from the preprocessing of ARRAU\textsuperscript{RST} to convert it into a format consistent with the \textsc{ICoref} model used for the other datasets in this work.}
    \label{tab:appendix:arrau}
\end{table}

\autoref{tab:data:train_size} shows the number of training examples we use for each dataset. Since we only shuffle once initially, larger training sets are always a superset of a smaller one. 

\begin{table}[]
    \centering
    \small
    \begin{tabular}{cl}
    \toprule
        Dataset & \# Training examples \\
        \midrule
         OntoNotes\textsuperscript{zh}  &  [0, 10, 25, 50, 100, 250, 500, 1810] \\
         OntoNotes\textsuperscript{ar}  &  [0, 10, 20, 40, 80, 160, 359] \\
         PreCo  & [5, 10, 25, 50, 100, 250, 500]\\
         LitBank & [5,10, 20, 40, 80] \\
         QBCoref & [5, 15, 30, 60, 120, 240] \\
         ARRAU\textsuperscript{RST}  & [10, 20, 40, 80, 160, 335]\\
         SARA & [10, 20, 40, 80, 138*] \\
         SemEval\textsuperscript{ca} & [10, 25, 50, 100, 250, 829] \\
         SemEval\textsuperscript{es} & [10, 25, 50, 100, 250, 875] \\
         SemEval\textsuperscript{it} & [10, 20, 40, 80] \\
         SemEval\textsuperscript{nl} & [10, 20, 40, 80, 145] \\
         \bottomrule
    \end{tabular}
    \caption{Training set sizes considered for each dataset. * For SARA, we use the entire fold, which contains 138 documents on average.}
    \label{tab:data:train_size}
\end{table}

\begin{figure*}[h]
    \centering
    \includegraphics[width=\linewidth]{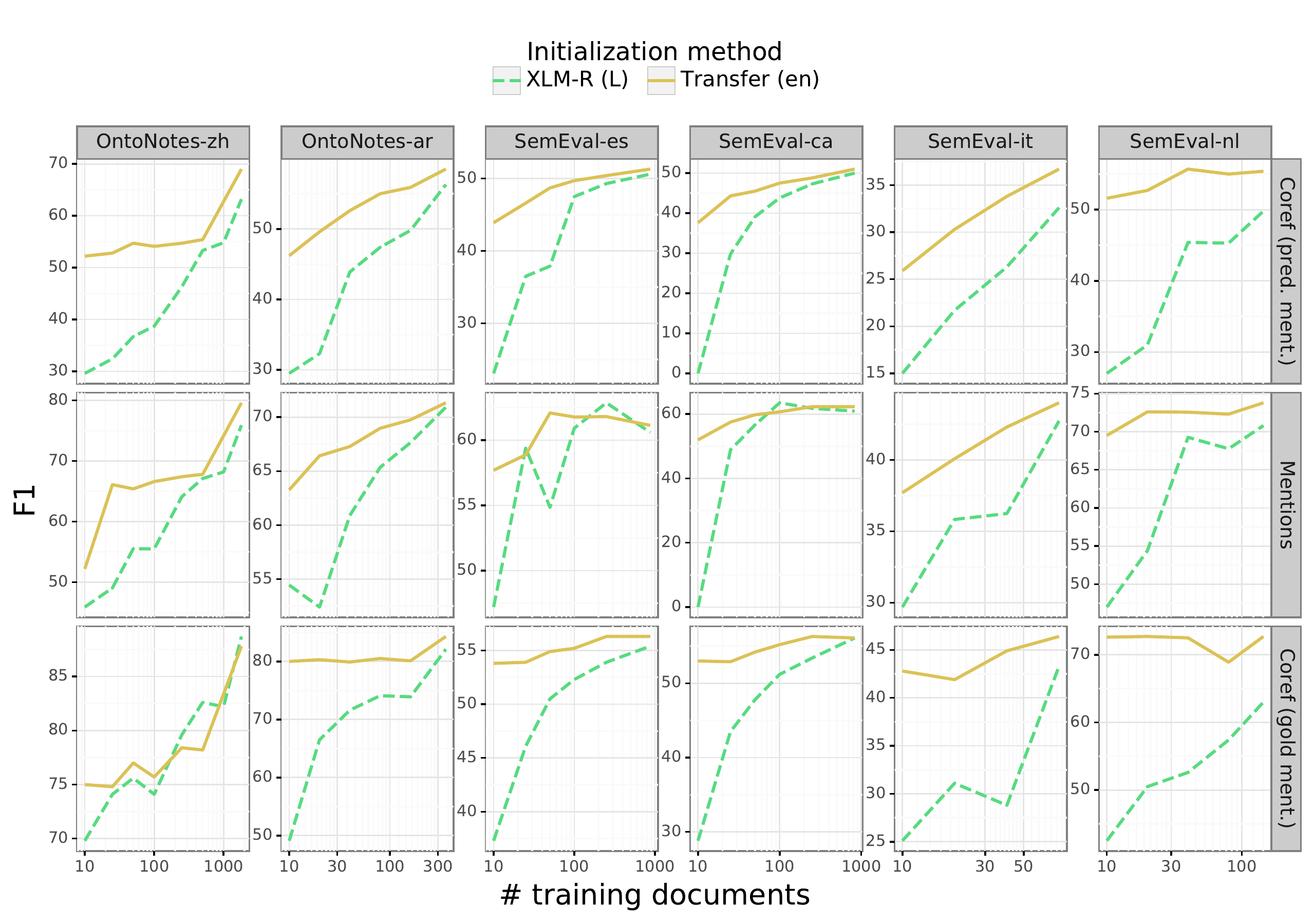}
    \caption{Full version of \autoref{fig:rq5links}. Like \autoref{fig:rq1links}, this plot demonstates the effectiveness of continued training across different \textit{languages}. \textsc{XLM-R} uses a pretrained encoder (dashed line), while \textsc{Transfer (en)} is first trained on OntoNotes\textsuperscript{en} (solid line).}
    \label{fig:rq5links_large}
\end{figure*}

\section{Training Details}
\label{sec:appendix:training}

We follow the same hyperparameters used by \citet{xia-etal-2020-incremental}. We use $k=0.4$ to select the top $0.4n$ spans, use learning rates of 2e-4 for training the non-encoder parameters (with Adam) and 1e-5 for the encoder (with AdamW). For all models, we finetune the full encoder. We use gradient clipping of 10, train for up to 100 epochs with a patience of 10 for early stopping, as determined by dev F1. We consider spans up to 10 for SARA, 15 for PreCo and ARRAU\textsuperscript{RST}, 20 for LitBank and QBCoref, and 30 for all other datasets. These choices are made based on prior work or the statistics of the training set; increasing the value would affect runtime (with marginal gains in performance).

Each model was trained on a single 24GB Nvidia Quadro RTX 6000s for between 20 minutes to 16 hours, depending on the number of training examples. Due to the cost of training over 500 models, each model was trained only once. The English models use 373M parameters, of which 334M is the SpanBERT-large encoder. The multilingual models use 599M parameters, of which 560M is XLM-R large. 

For Table \ref{tab:data:benchmark}, we pick the best model between \textsc{Transfer (on)}  and \textsc{Transfer (pc)} based on their dev scores on each dataset. These are listed in \autoref{tab:appendix:dev}

\begin{table}[h]
    \centering
    \small
    \begin{tabular}{cccc}
    \toprule
         Dataset & \textsc{on} & \textsc{pc} & \textsc{en} \\
    \midrule
         PreCo & 82.4 &  \textbf{85.2} & -\\
         LitBank & \textbf{77.3} & 76.3 & -\\
         QBCoref & \textbf{79.1} & 78.7 & - \\
         ARRAU\textsuperscript{RST} & 77.7 & \textbf{79.3} & -\\
         SARA & \textbf{77.7} & 75.4 & -\\
    \midrule
        OntoNotes\textsuperscript{zh} & - & - & \textbf{69.0} \\
        OntoNotes\textsuperscript{ar} & - & - & \textbf{62.3} \\
        SemEval\textsuperscript{ca} & - & - & \textbf{51.4} \\
        SemEval\textsuperscript{es} & - & - & \textbf{52.1} \\
        SemEval\textsuperscript{it} & - & - & \textbf{36.1} \\
        SemEval\textsuperscript{nl} & - & - & \textbf{48.3}\\
    \bottomrule
    \end{tabular}
    \caption{Dev. F1 scores on each of the models and datasets presented in \autoref{tab:data:benchmark}. For the English dataset, the test score of the model with the best performing score is reported in \autoref{tab:data:benchmark}.}
    \label{tab:appendix:dev}
\end{table}

\section{Full cross-lingual results}
\label{sec:appendix:layers}

\subsection{Continued Training (\ref{sec:res:one})}

\autoref{fig:rq5links_large} is the full figure, analogous to \autoref{fig:rq1links}.

\begin{figure*}
    \centering
    \includegraphics[width=0.7\linewidth]{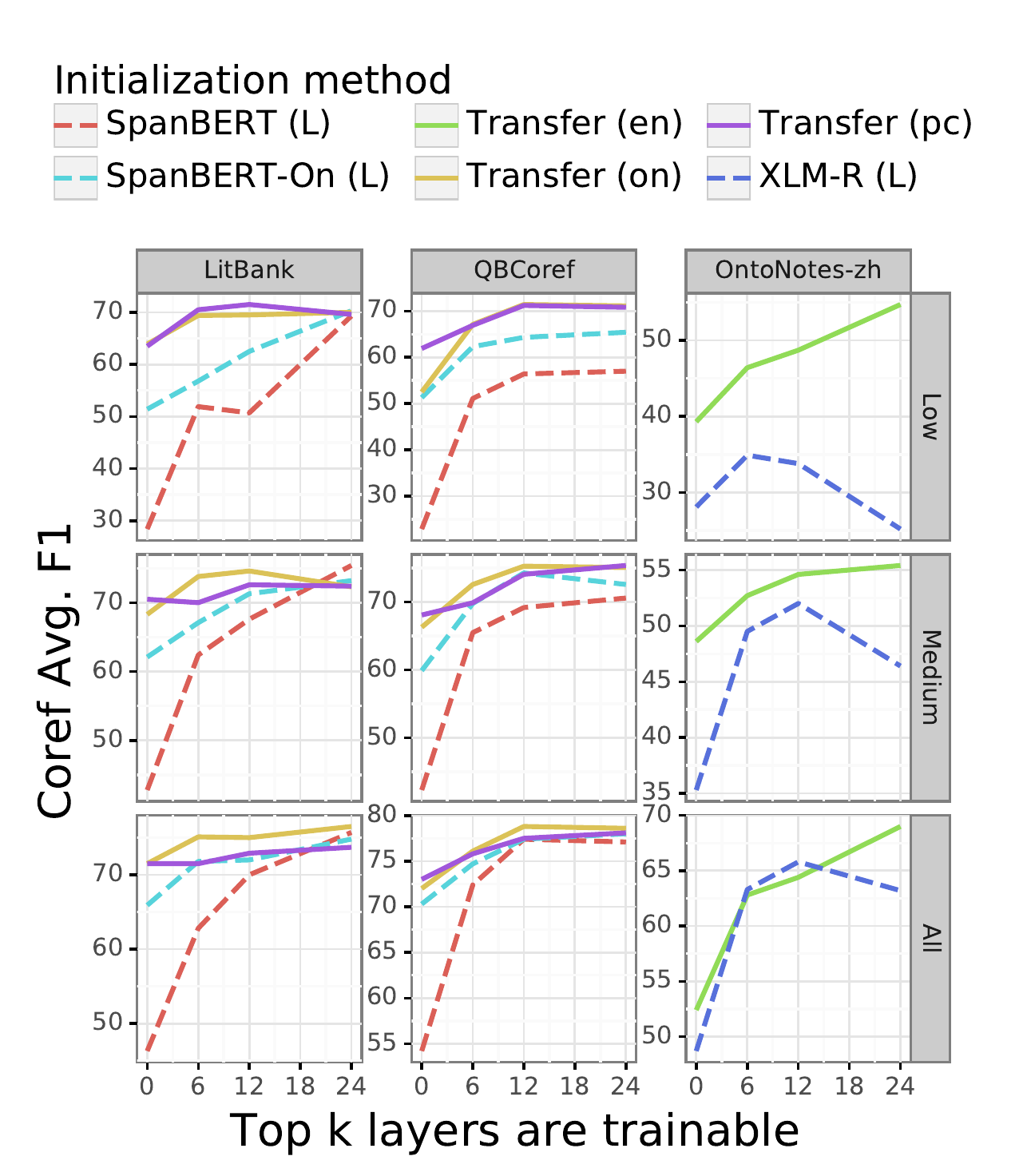}
    \caption{Average F\textsubscript{1} across different models and number of trainable layers, varying between 0, 6, 12 or 24 layers. \textit{Low} vs. \textit{Medium} vs. \textit{All} describes the number of documents used for the first fold of LitBank (10, 40, 80 documents), QBCoref (15, 60, 240 documents), and OntoNotes\textsuperscript{zh} (50, 500, 1810 documents). The initialization methods follow those used throughout the paper.}
    \label{fig:rq4_full}
\end{figure*}

\subsection{Training Top Layers (\ref{sec:res:layers})}
For each dataset, we include a "medium" data volume (40 for LitBank, 60 for QBCoref) and we also include OntoNotes\textsuperscript{zh} with 50, 500, and 1810 for the three data volumes respectively. These plots are presented in \autoref{fig:rq4_full}. These trends follow what is described in \autoref{sec:res:layers}. Notably, freezing the lower layers when training OntoNotes\textsuperscript{zh} from scratch appears to consistently outperform training the full model.

\end{document}